\documentclass{bmvc2k}
\usepackage{mathrsfs}
\usepackage{bm}
\usepackage{amsmath}
\usepackage{amssymb}
\usepackage{booktabs}
\usepackage{array}
\usepackage{graphics}

\title{Query-Conditioned Three-Player Adversarial Network for Video Summarization}

\addauthor{Yujia Zhang}{zhangyujia2014@ia.ac.cn}{12}
\addauthor{Michael Kampffmeyer}{michael.c.kampffmeyer@uit.no}{3}
\addauthor{Xiaodan Liang}{xdliang328@gmail.com}{4}
\addauthor{Min Tan}{min.tan@ia.ac.cn}{12}
\addauthor{Eric P. Xing}{epxing@cs.cmu.edu}{4}

\addinstitution{
Institute of Automation,\\
Chinese Academy of Sciences
}
\addinstitution{
University of Chinese Academy of \\Sciences
}
\addinstitution{
UiT The Arctic University of Norway
}
\addinstitution{
Carnegie Mellon University
}

\runninghead{ZHANG et al.}{QUERY-CONDITIONED VIDEO SUMMARIZATION}

\renewcommand{\paragraph}[1]{\vspace{0.2cm}\noindent{\bf #1}}

\begin{document}
\newcolumntype{?}{!{\vrule width 0.5pt}!{\vrule width 0.5pt}}
\maketitle

\begin{abstract}
Video summarization plays an important role in video understanding by selecting key frames/shots. Traditionally, it aims to find the most representative and diverse contents in a video as short summaries. Recently, a more generalized task, query-conditioned video summarization, has been introduced, which takes user queries into consideration to learn more user-oriented summaries. In this paper, we propose a query-conditioned three-player generative adversarial network to tackle this challenge. The generator learns the joint representation of the user query and the video content, and the discriminator takes three pairs of query-conditioned summaries as the input to discriminate the real summary from a generated and a random one. A three-player loss is introduced for joint training of the generator and the discriminator, which forces the generator to learn better summary results, and avoids the generation of random trivial summaries. Experiments on a recently proposed query-conditioned video summarization benchmark dataset show the efficiency and efficacy of our proposed method.
\end{abstract}

\section{Introduction}
\label{sec:intro}

\begin{figure*}[t]
\centerline{\includegraphics[width=\textwidth]{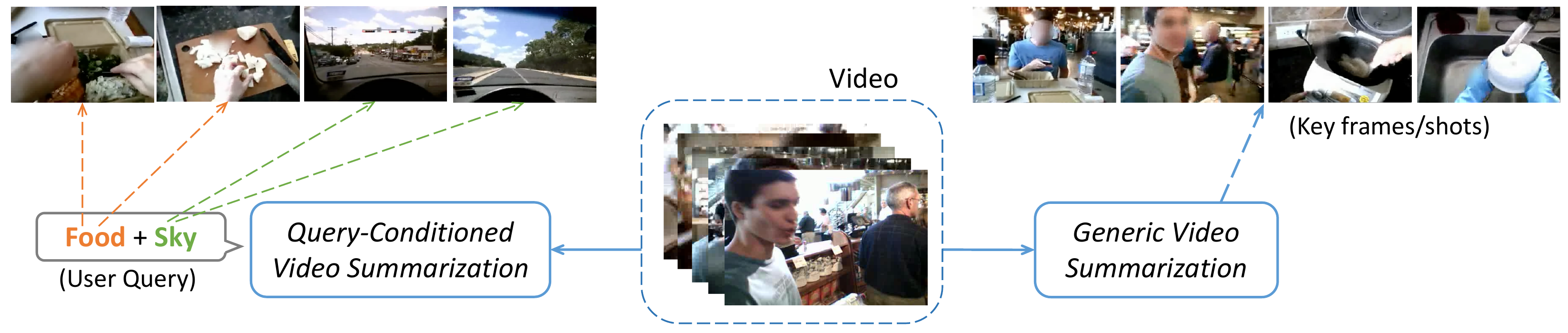}}
\caption{Different video summarization tasks. Generic video summarization aims to generate key contents of a video, while query-conditioned video summarization takes the user query into consideration and generates summaries accordingly.}
\label{fig:generic_query}
\end{figure*}

Video summarization aims to select key frames/shots among videos to summarize the main storyline and has been widely investigated for facilitating video understanding~\cite{plummer2017enhancing,mahasseni2017unsupervised,zhou2017deep,yuan2017temporal,han2018reinforcement,goyal2017nonparametric}. As shown in Figure~\ref{fig:generic_query}, this task can be classified into two types: a) generic video summarization, which only takes the visual features of the video contents as the input and b) query-conditioned video summarization which conditions summarization on user queries.

The generic video summarization task has been addressed at three different levels: shot-level~\cite{lin2015summarizing,lu2013story}, frame-level~\cite{kim2014joint,khosla2013large}, and object-level~\cite{meng2016keyframes,zhang2018unsupervised} video summarization by selecting key shots/frames/objects in the videos. However, one main issue with generic video summarization is the fact that it does not take user preferences into account, since different users may have different preferences towards the video content, and a single evaluation metric is not robust enough for all video summaries~\cite{sharghi2017query}.

Recently, another research direction, query-conditioned video summarization~\cite{sharghi2016query,vasudevan2017query,sharghi2017query}, has been explored, which takes advantage of different user queries in form of texts to learn more user-oriented summaries. It generates user-oriented summaries that have effective correlations between summaries and query, and capture the overall essence of the video. Several approaches to query-conditioned video summarization have been proposed. Sharghi et al.~\cite{sharghi2016query} first extend a sequential DPP (seqDPP)~\cite{gong2014diverse} to extract key shots. Afterwards, they develop a more comprehensive dataset for this task, and propose a memory network~\cite{sukhbaatar2015end} parameterized seqDPP model. However, there is still room to learn a better summarizer due to the limitation of the memory to jointly encode video and query.

To address the above issue we develop a query-conditioned three-player generative adversarial network architecture. We encode the query and the video sequence to learn a joint representation combining visual and text information, and take this query-conditioned representation as the input to the generative adversarial network. A three-player structure is applied during joint training, in order to achieve superior regularization.
The contribution in our work can be summarized as follows: first, we propose a query-conditioned three-player adversarial network, which jointly encodes query and visual information and learns in an adversarial manner. Second, we introduce a three-player structure for the adversarial training. The discriminator regularizes the model via the three-player loss, which facilitates the generator to generate more related and meaningful video summaries. Two supervised losses are applied to ensure a more compact summary. One loss regularizes the length and the other aligns prediction and ground truth. Experimental results on a public dataset~\cite{sharghi2017query} demonstrate the superiority of our proposed approach against the state-of-the-art method.

\section{Related Work}

\subsection{Generic Video Summarization}
Generic video summarization~\cite{kim2014joint,khosla2013large,zhang2016summary,zhang2016video,lin2015summarizing}, has been widely studied for efficient video analysis and video understanding.
For shot-level video summarization~\cite{lu2013story,yao2016highlight,song2015tvsum,lin2015summarizing}, Song et al.~\cite{song2015tvsum} propose to learn the canonical visual concepts which are shared between videos and images to find important shots. In~\cite{yao2016highlight}, a pairwise deep ranking model is proposed to distinguish highlight segments from non-highlight ones. 
For frame-level video summarization~\cite{khosla2013large,kim2014joint,gong2014diverse,zhang2018dtr}, Khosla et al.~\cite{khosla2013large} use web-images as a prior to facilitate video summarization. In~\cite{gong2014diverse}, a probabilistic model is proposed for learning sequential structures to generate summaries.
Approaches to object-level video summarization~\cite{meng2016keyframes,zhang2018unsupervised} aim to obtain representative objects to perform fine-grained summarization. Currently there are two existing GAN-based works~\cite{mahasseni2017unsupervised,zhang2018dtr} that include regularization using adversarial training. However, they do not consider user preferences, so the summaries may not be robust and may not generalize well to different users. Therefore, we investigate the query-conditioned video summarization task to provide more personalized summarization results by relying on user queries.

\subsection{Query-conditioned Video Summarization}
Query-conditioned video summarization~\cite{sharghi2016query,sharghi2017query,vasudevan2017query,oosterhuis2016semantic,ji2017query} takes user queries in the form of texts into consideration in order to learn more user-oriented summaries. In~\cite{sharghi2016query}, a Sequential and Hierarchical DPP (SH-DPP) is developed to tackle this challenge. In~\cite{vasudevan2017query}, the authors adopt a quality-aware relevance model and submodular mixtures to pick relevant and representative frames. There are two other works related to query-conditioned video summarization. One is used to generate visual trailers, while the other obtains web images conditioned on user queries, and then produces video summaries from both images and videos. Specifically, Oosterhuis et al.~\cite{oosterhuis2016semantic} propose a graph-based method to generate visual trailers by selecting frames that are most relevant to a given user's query. Ji et al.~\cite{ji2017query} formulate the task by incorporating web images, which are obtained from user query searches. Thus the video summarization is indirectly conditioned on the query through the web images.

Recently, Sharghi et al.~\cite{sharghi2017query} explore a more thoroughly query-conditioned video summarization approach. Instead of using datasets which are originally collected for the generic task, they propose a new dataset and an evaluation metric towards this task. Our work is developed based on this new dataset and the evaluation metric. We propose a novel query-conditioned adversarial network which does not rely on external knowledge, such as web images, and can effectively summarize videos based on user queries by integrating a three-player adversarial training structure.

\section{Proposed Algorithm}
\subsection{Generator Network}
\label{generator}

\begin{figure*}[t]
    \centering
    \includegraphics[width=\textwidth]{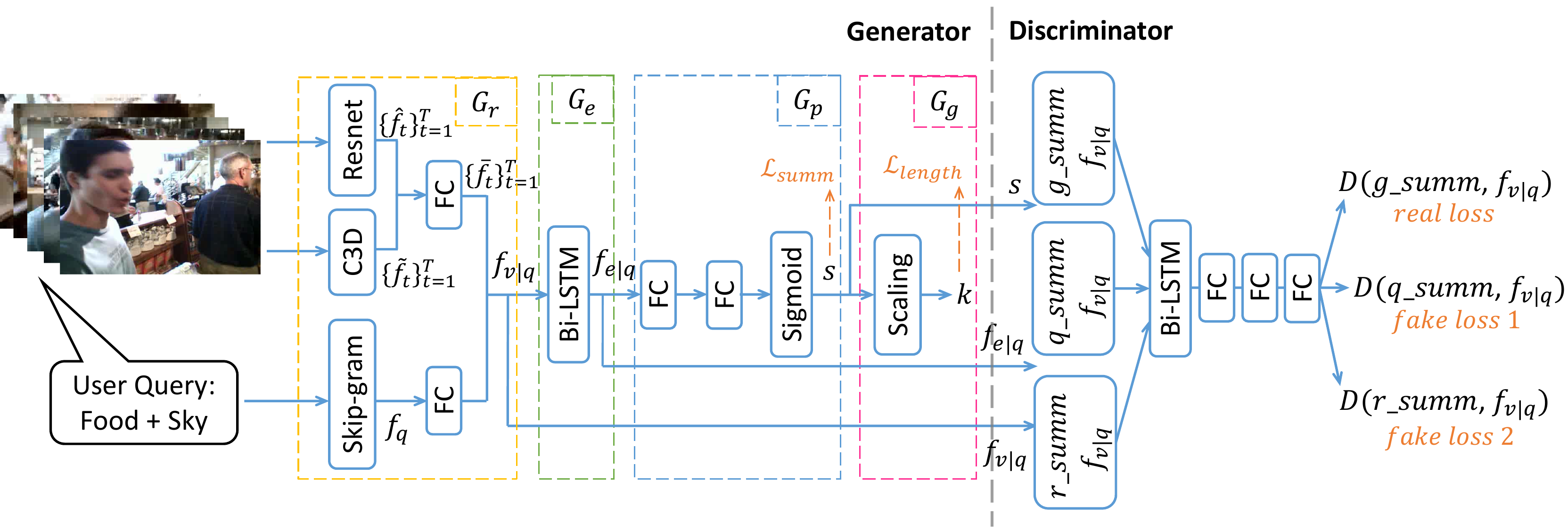}
    \caption{The network architecture of our proposed method for query-conditioned video summarization. In the generator, the video is fed into a query-conditioned feature representation module $G_r$, that integrates query and visual information. After a compact video encoding process in module $G_e$, we can predict shot scores in module $G_p$. The summary is then generated using video summary generator $G_g$, with the final results. We further introduce two regularizations: the summary regularization $\mathcal{L}_{summ}$ and the length regularization $\mathcal{L}_{length}$ to enhance the generator's ability to learn superior summaries. The discriminator uses three query-conditioned representations as the input, and is tasked to distinguish the real summary from two fake summaries in an adversarial learning manner.}
    \label{framework}
\end{figure*}

Our proposed network facilitates query-conditioned video summarization by applying an adversarial network that takes the query into consideration with a three-player discriminator loss. Figure~\ref{framework} illustrates the whole framework of our approach. The generator is mainly tasked with embedding visual information and text jointly, in order to provide comprehensive query-conditioned representations. The discriminator aims to distinguish the real summaries, i.e., ground-truth summary from random and generated summaries.

In the following sections, we first introduce the query-conditioned generator network in our model to select key shots with regards to different queries. We then present the proposed query-conditioned discriminator with three-player loss, which distinguishes the ground-truth summary from the random and the generated summaries. Finally, we introduce the details of adversarial training with two supervision losses in our model.

\subsubsection{Query-Conditioned Feature Representation Module $G_r$}
\paragraph{Frame-level visual representation.} We denote an input video as $\mathcal{V}=\{\mathcal{V}_t\}_{t=1}^T$, where $T$ denotes the total number of shots in a video. Each video shot $\{\mathcal{V}_t\}$ is partitioned into 75 frames (5-second long) for fair comparison with related work~\cite{sharghi2017query}. As shown in Figure~\ref{framework}, the model $G_r$ aims to generate feature representations which are conditioned on user queries. We first apply the ResNet-152 feature extractor~\cite{he2016deep} to encode frame-level visual features. In order to do this, we downsample each shot to 16 frames per segment. The ``fc7'' layer of the ResNet-152 model trained on the ILSVRC 2015 dataset~\cite{russakovsky2015imagenet} is used to obtain features for frames within each shot and followed by an average pooling layer. This frame-level feature vector is denoted as $\{\hat{f_t}\}_{t=1}^T$.

\paragraph{Shot-level visual representation.} We apply the C3D video descriptor~\cite{tran2015learning}, a network trained on the Sports1M dataset~\cite{tran2015learning}, to extract shot-level feature representation. We use the output of the ``fc6'' layer of the C3D and uniformly split the video into 75 frames before downsampling it to 16 frames per segment, aligning it with the extracted ResNet features, to get the shot-level visual features. The features we extracted from the C3D are denoted as $\{\Tilde{f_t}\}_{t=1}^T$. 

\paragraph{Textual representation.} To obtain textual feature representations, we use the Skip-gram model~\cite{mikolov2013distributed}, a word2vec model pretrained on the GoogleNews dataset, to project each word into a semantic feature space. We define each user query as $q$. Each $q$ contains two concepts (words), and we generate the concept embedding by summing up the two feature vectors of the two concepts. After that, we encode the textual representation $f_q$ by applying a fully connected layer. 

\paragraph{Query-Conditioned Feature representation.} We first combine frame-level and shot-level feature vectors $\{\hat{f_t}\}_{t=1}^T$ and $\{\Tilde{f_t}\}_{t=1}^T$ by concatenation, followed by a fully connected layer to get the encoded visual feature $\{\Bar{f_t}\}_{t=1}^T$. After that, we use another concatenation to combine $\{\Bar{f_t}\}_{t=1}^T$ and the textual representation $f_q$. Thus, we obtain a query-conditioned feature encoding for the video, which can be denoted as $f_{v|q}=\{f_1^v,f_2^v, \dots ,f_T^v|q\}$, i.e., $f_{v|q}=G_r(\mathcal{V}|q)$.

\subsubsection{Video Summarization Prediction}
\paragraph{Compact Video Encoding Module $G_e$.}
Given the generated query-conditioned feature representation $f_v$ from model $G_r$, we introduce the compact video encoding module $G_e$ to learn the temporal dependencies among video shots. Thus the output of the compact video encoding can be produced as $f_{e|q}=G_e(f_{v|q})$, where $f_{e|q}=\{f_1^e,f_2^e, \dots ,f_T^e|q\}$. The model $G_e$ consists of a Bidirectional LSTM (Bi-LSTM) layer~\cite{graves2005framewise} to model the temporal representation, followed by a batch normalization layer~\cite{ioffe2015batch} and Rectified Linear Unit (ReLU) activation~\cite{glorot2011deep} to learn the compact video encoding.

\paragraph{Shot Score Prediction Module $G_p$.}
In order to predict a confidence score for each video shot, we propose the shot score prediction module $G_p$. We define the confidence score as $s=\{s_t\}_{t=1}^T$, where $s_t=G_p(f_t^e|q)$. In our setting, we use two fully connected layers with a batch normalization and a ReLU activation in the middle. After that, we apply a sigmoid layer in order to generate a confidence score for each video shot being a key shot. 

\paragraph{Video Summary Generator $G_g$.}
Given the confidence score of each video shot from model $G_p$, we introduce the video summary generator $G_g$ to generate the final results for selected key shots by means of scaling. To get the video summary results, we apply $k=G_g(s)$ by passing the shot score into the video summary generator, where $k$ is the summary result for the shot, and $k=\{k_t\}_{t=1}^T$. $k_t=0$ means that the shot is a trivial one, while $k_t=1$ represents that it is a key shot which will be included in the generated summary. $G_g$ is a softmax function with a temperature parameter $\tau$ to get the result of each $k_t$:

\begin{equation}
    k_t=\frac{e^{s_t/\tau}}{e^{s_t/\tau}+e^{(1-s_t)/\tau}}.
\end{equation}

\subsection{Discriminator Network}
\label{discriminator}
We use three pairs of different summaries together with the feature representation of the video as the input to the discriminator. For simplicity, we use $q_{summ}$, $g_{summ}$ and $r_{summ}$ to denote \emph{generated query-conditioned summary}, \emph{ground-truth query-conditioned summary}, and \emph{randomly generated query-conditioned summary}, respectively. The three pairs are: \emph{($q_{summ}$, video shots)}, \emph{($g_{summ}$, video shots)}, and \emph{($r_{summ}$, video shots)}. The \emph{video shots} we use are the learn joint embedding of visual and query information. We use $r_{summ}$ to enhance the ability of the generator to learn a more robust summary as well as avoid the generation of random trivial short sequences.

As shown in Figure~\ref{framework}, we use query-conditioned feature representation $f_{v|q}$ generated from model $G_R$ as the input of \emph{video shots} in the three pairs as the learned feature for the video. $r_{summ}$ is obtained using the random summary score $s_r$ and by generating random values of 0 and 1. The length of $s_r$ is the same as the one of predicted summary $s$ from the video summary generator $G_g$. $g_{summ}$ is produced using a ground-truth summary score $s_g$, where $s_g=\{s_1^g,s_2^g,\dots,s_T^g\}$. Note, $s_r$,$s_g$ $\in [0,1]$. In order to get $(q_{summ}, f_{v|q})$, $(g_{summ}, f_{v|q})$ and $(r_{summ}, f_{v|q})$, the three summary representations can be defined as:

\begin{align}
    \begin{split}
    & q_{summ}=f_{e|q}\cdot s, \\
    & g_{summ}=f_{e|q}\cdot s_g, \\
    & r_{summ}=f_{e|q}\cdot s_r. \\
    \end{split}
\end{align}
After that, we take $f_{v|q}$ as the input to a Bi-LSTM layer, followed by a batch normalization layer and ReLU activation, and pass $q_{summ}$, $g_{summ}$ and $r_{summ}$ to another Bi-LSTM and a batch normalization layer with ReLU activation to learn a temporal representation. Then we concatenate them in pairs and apply three fully connected layers and jointly train the discriminator to distinguish the true summary from fake ones.

\subsection{Adversarial Learning}
We first introduce the summary regularization $\mathcal{L}_{summ}$ to optimize the generator by enforcing the selection of key shots to align with the ground-truth. It aligns a predicted shot score $s_t$ from the model $G_p$ with the corresponding ground-truth summary score $s_t^g$: 

\begin{equation}
    \mathcal{L}_{summ}=\frac{1}{T}\sum_{t=1}^{T}(s_t-s_t^g)^2.
\end{equation}  
We further incorporate the length regularization $\mathcal{L}_{length}$, which is computed between the number of generated summary shots and the ground-truth summary during the adversarial training to control the length of summaries:

\begin{equation}
    \mathcal{L}_{length}=\left|\frac{1}{T}\sum_{t=1}^{T}k_t-\gamma\right|,
\end{equation}
where $\gamma$ is the percentage of the key shots in the the video based on ground-truth summary, and $k_t$ is the summary result for each video shot generated from the model $G_g$.

Our adversarial objective function is based on Wasserstein GANs~\cite{arjovsky2017wasserstein}, due to its good convergence property. Note that this does not exclude the use of other GAN-based objectives, as our model is flexible enough to be combined with other GAN structures.

Instead of using the commonly used two-player learning mode, we optimize it with the three-player loss as shown in Figure~\ref{framework}: the real loss $D(\emph{g\_{summ}},f\_{v|q})$ and the two fake losses $D(\emph{q\_{summ}},f\_{v|q})$ and $D(\emph{r\_{summ}},f\_{v|q})$. The thee-player loss can not only force the model to generate good summaries, but can also avoid the learning of a trivial summary of randomly selected shots.

The generator $G$ and the discriminator $D$ conditioned on query $q$ are jointly optimized by the use of a min-max adversarial objective:

\begin{align}
    \begin{split}
    \underset{G}{\text{min}}&\underset{D}{\text{max}}\mathcal{L}(G,D|q) = {\mathbb{E}_g}[D(\emph{$g_{summ}$},f_{v|q})]\\&
    -\omega{\mathbb{E}_q}[D(\emph{$q_{summ}$},f_{v|q})] -(1-\omega){\mathbb{E}_r}[D(\emph{$r_{summ}$},f_{v|q})],
    \end{split}
\end{align}
where $\omega$ is the balancing parameter for the two fake losses. Here we use $\omega=0.5$ for treating the two fake losses equally. We replace the generator $G$ with models $G_r$, $G_e$, $G_p$ in Section~\ref{generator} and the three summary representation $q_{summ}$, $g_{summ}$ and $r_{summ}$ in Section~\ref{discriminator}, so that the objective function Eq. (5) can be reformulated as:

\begin{align}
    \begin{split}
    & \underset{G_r,G_e,G_p}{\text{min}}\underset{D}{\text{max}}\mathcal{L}(G_r,G_e,G_p,D|q)= \\ 
    & {\mathbb{E}_g}[D(G_e(G_r(\mathcal{V}|q))\cdot s_g,G_r(\mathcal{V}|q)] \\
    & -0.5{\mathbb{E}_q}[D(G_e(G_r(\mathcal{V}|q))\cdot G_p(G_e(G_r(\mathcal{V}|q)),G_r(\mathcal{V}|q)] \\
    & -0.5{\mathbb{E}_r}[D(G_e(G_r(\mathcal{V}|q))\cdot s_r,),G_r(\mathcal{V}|q)].
    \end{split}
\end{align}
Thus, the final objective function conditioned on query $q$ including the two supervised losses, $\mathcal{L}_{summ}$ and $\mathcal{L}_{length}$, can be denoted as $G^*$:

\begin{equation}
    G^*=\text{arg}\underset{G_r,G_e,G_p}{\text{min}}\underset{D}{\text{max}}\mathcal{L}(G_r,G_e,G_p,D|q)+\mathcal{L}_{summ}+\mathcal{L}_{length}\;.
\end{equation}

\section{Experimental Results}
\subsection{Datasets and Settings}
\paragraph{Datasets.}
We evaluate our approach on the query-conditioned dataset proposed in~\cite{sharghi2017query}, which is built upon the existing UT Egocentric (UTE) dataset~\cite{lee2012discovering}. The dataset has 4 videos in total, containing different uncontrolled daily lives scenarios and each being 3$\sim$5 hours long. A dictionary of concepts for user queries is supplied, which is a concise and diverse set of 48 concepts, which are deemed to be comprehensive of daily life for the query-conditioned video summarization. As for the queries, four different scenarios are included to formalize comprehensive queries~\cite{sharghi2017query}. Note that among different queries, one scenario is introduced where none of the concepts in the query are presented in the video. The three remaining scenarios are 1) queries, where all concepts appear together in the same video shot, 2) queries, where all concepts appear but not in the same shot, and 3) queries, where only one of the concepts appears.
For fair comparison, we follow~\cite{sharghi2017query} and randomly select two videos for training, leaving one for testing and one for validation. Four experiments are performed to test all four videos.

\paragraph{Evaluation Metrics.}
In~\cite{sharghi2017query}, the authors propose to find the ideal mapping between the generated query and the ground-truths summary by the maximum weight matching of a bipartite graph, based on a similarity function between two video shots. The similarity function uses the intersection-over-onion (IoU) on corresponding concepts to evaluate the performance. The IoU is defined as the edge weights, and the generated and ground-truths summaries are on opposing sides of the graph. 
Precision, recall, and F1-score are computed based on the number of matched summary pairs.

\subsection{Implementation Details}
We implement our work using TensorFlow~\cite{abadi2016tensorflow}, with 1 GTX TITAN X 12GB card on a single server. In the generator $G$, the output for frame- and shot-level visual representation are 2048- and 4096-dimensional vectors, and the textual representation is a 300-dimensional vector. The learned temporal representation after the Bi-LSTM is 2048-dimension. In the module $G_p$, the output of the two fully connected layers are 128- and 1-dimensional vectors, respectively. We use a low-temperature softmax function in module $G_g$ in order to get approximately binary results. The dropout between the two fully connected layers is 0.5.

In the discriminator $D$, the Bi-LSTM that we use encodes the features to a 512-dimensional vector. The output dimensions for the three fully connected layers are 512-, 256-, and 128-dimensions. During the training phase, we randomly select a set of 1000 successive video shots for each batch, with one user query for all shots. For the generic scenario where none of the concepts in the query are presented in the video, we use a zero vector of 300-dimension for the query embedding. During the testing phase, we obtain the predicted shot score in module $G_p$ for each video shot.

The inference times for the 4 videos are 1472$ms$, 1948$ms$, 1141$ms$ and 1893$ms$, respectively, so on average, it takes 1.614$s$ for each video to generate query-conditioned key video shots.

\subsection{Quantitative Results}
\subsubsection{Comparison Analysis}
We compare our approach with all other frameworks which have been applied to this query-conditioned video summarization dataset. The precision, recall and F1-score comparison for the four videos are shown in Table~\ref{tab:results}. It can be observed that our approach outperforms the existing state-of-the-art by 1.86\%. Especially for Video 2 and Video 4, we achieve 3.64\% and 3.68\% better performance than~\cite{sharghi2017query} in terms of F1-score. Such substantial performance improvements indicate the superiority of our proposed method by using a three-player adversarial network on the joint embedding of visual information and the user query. The rest three works are all based on a DPP architecture, which can learn long time temporal relations among video shots. However, our work adopts the adversarial learning objective, which facilitates both temporal and query-conditioned joint learning. The two regularizations on summary and length also help obtaining better query-conditioned summary.

\bgroup
\def\arraystretch{1}
\setlength{\tabcolsep}{4pt}
\begin{table}[tbp] \small
\centering
\caption{Results obtained by our method compared to other approaches for query-conditioned video summarization in terms of Precision (Pre), Recall (Rec) and F1-score(F1).}
\label{tab:results}
\begin{tabular}{l?ccc?ccc?ccc?ccc}
\toprule
\multicolumn{1}{c?}{}& \multicolumn{3}{c?}{SeqDPP~\cite{gong2014diverse}} & \multicolumn{3}{c?}{SH-DPP~\cite{sharghi2016query}} & \multicolumn{3}{c?}{QC-DPP~\cite{sharghi2017query}} & \multicolumn{3}{c}{\textbf{Ours}}\\
\cmidrule(lr){2-4}
\cmidrule(lr){5-7}
\cmidrule(lr){8-10}
\cmidrule(lr){11-13}
& Pre & Rec & F1 & Pre & Rec & F1 & Pre & Rec & F1 & Pre & Rec & F1\\
\midrule
Vid1 & 53.43 & 29.81 & 36.59 & 50.56 & 29.64 & 35.67 & 49.86 & 53.38 & 48.68 & 49.66 & 50.91 & \textbf{48.74}\\
Vid2 & 44.05 & 46.65 & 43.67 & 42.13 & 46.81 & 42.72 & 33.71 & 62.09 & 41.66 & 43.02 & 48.73 & \textbf{45.30}\\
Vid3 & 49.25 & 17.44 & 25.26 & 51.92 & 29.24 & 36.51 & 55.16 & 62.40 & 56.47 & 58.73 & 56.49 & \textbf{56.51} \\
Vid4 & 11.14 & 63.49 & 18.15 & 11.51 & 62.88 & 18.62 & 21.39 & 63.12 & 29.96 & 36.70 & 35.96 & \textbf{33.64}\\
\midrule
Avg. & 39.47 & 39.35 & 30.92 & 39.03 & 42.14 & 33.38 & 40.03 & 60.25 & 44.19 & 47.03 & 48.02 & \textbf{46.05} \\
\bottomrule
\end{tabular}
\end{table}
\egroup

\subsubsection{Ablation Analysis}

\begin{table}[h]  \small
    \centering
    \renewcommand\arraystretch{1.2}
    \caption{Ablation analysis on query-conditioned video summarization in terms of Precision (Pre), Recall (Rec) and F1-score (F1).}\label{ablation}
    \begin{tabular}{c?c|c|c}
        \toprule
        Method & Pre & Rec & F1 \\
        \midrule
        \textbf{Ours} & 43.02 & 48.73 & \textbf{45.30} \\
        \emph{w/o-$\mathcal{L}_{length}$} & 34.78 & 61.80 & 44.08 \\
        \emph{w/o-$\mathcal{L}_{summ}$} & 28.30 & 47.58 & 35.19 \\
        \emph{two-player} & 43.39 & 51.28 & 44.37 \\
        \bottomrule
    \end{tabular}
\end{table}

\bgroup
\def\arraystretch{1}
\setlength{\tabcolsep}{4pt}
\begin{table}[tbp] \small
\centering
\caption{Query length analysis on query-conditioned video summarization in terms of Precision (Pre), Recall (Rec) and F1-score (F1).}\label{ablation2}
\begin{tabular}{l?c|ccc?c|ccc}
\toprule
\multicolumn{1}{c?}{}& \multicolumn{4}{c?}{\emph{w/o-$\mathcal{L}_{length}$}} & \multicolumn{4}{c}{\textbf{Ours} \emph{(w-$\mathcal{L}_{length}$)}} \\
\cmidrule(lr){2-5}
\cmidrule(lr){6-9}
& $d_{w/o-len}$ & Pre & Rec & F1 & $d_{w-len}$ & Pre & Rec & F1 \\
\midrule
Vid1 & 50.72 & 45.42 & 57.09 & 47.45 & \textbf{26.09} & 49.66 & 50.91 & \textbf{48.74} \\
Vid2 & 50.23 & 34.78 & 61.80 & 44.08 & \textbf{11.59} & 43.02 & 48.73 & \textbf{45.30} \\
Vid3 & 31.15 & 48.50 & 63.59 & 52.98 & \textbf{13.61} & 58.73 & 56.49 & \textbf{56.51} \\
Vid4 & 44.67 & 23.46 & 51.96 & 31.69 & \textbf{17.98} & 36.70 & 35.96 & \textbf{33.64} \\
\midrule
Avg. & 40.88 & 40.19 & 55.98 & 44.12 & \textbf{17.32} & 47.03 & 48.02 & \textbf{46.05} \\
\bottomrule
\end{tabular}
\end{table}
\egroup

We conduct experiments on different components of our model. As shown in Table~\ref{ablation}, we use \emph{$w/o-\mathcal{L}_{length}$}, \emph{$w/o-\mathcal{L}_{summ}$} and \emph{$two-player$} to denote our model when trained without the length regularization loss, the ground-truth summary regularization loss, and the random summary loss respectively. We can observe that the performance is reduced slightly after dropping the length regularization and the random summary as a form of two-player structure. Thus it demonstrates the effects of the length regularization and the three-player manner. Besides, there is a large decline after dropping the ground-truth summary regularization, which complies with the fact that additional supervised information tends to improve learning considerably.

We further conduct an experiment to more thoroughly analyze the ability of our proposed summary length regularization approach to generate summaries of suitable length. Here we define the summary length distance between the generated summary and the ground-truth summary as: $d_{w-len}=\left|\frac{1}{Q}\sum_{q=1}^{Q}\sum_{t=1}^{T}(k_{t,q}-s_{t,q}^g)\right|$, $Q$ is the total number of queries. We use $k_{t,q}$ and $s_{t,q}^g$ to denote the summary result and the ground-truth summary given a certain query $q$ with the length regularization $\mathcal{L}_{length}$. Similarly, the summary distance between the generated summary after dropping the length regularization and the ground-truth summary is defined as: $d_{w/o-len}=\left|\frac{1}{Q}\sum_{q=1}^{Q}\sum_{t=1}^{T}(k_{t,q}'-s_{t,q}^g)\right|$, $k_{t,q}'$ denotes the summary result given a certain query $q$ after dropping the length regularization $\mathcal{L}_{length}$. 

As shown in Table~\ref{ablation2}, the F1-score of the model without length regularization is reduced by 1.93\% in average, compared with the result of our proposed framework, and the length distance increases from 17.32 to 40.88. This demonstrates the effect of the length regularization. Moreover, we can also observe that the differences between precision and recall values for \emph{$w/o-\mathcal{L}_{length}$} tend to be larger than the ones in our proposed approach. Note that the smaller the distance between precision and recall values is, the closer between the length of the summary and the length of the ground-truth is. This indicates the effects of our introduced query length regularization $\mathcal{L}_{length}$.

\subsection{Qualitative Results}
\begin{figure*}[t]
    \centering
    \includegraphics[width=0.9\textwidth]{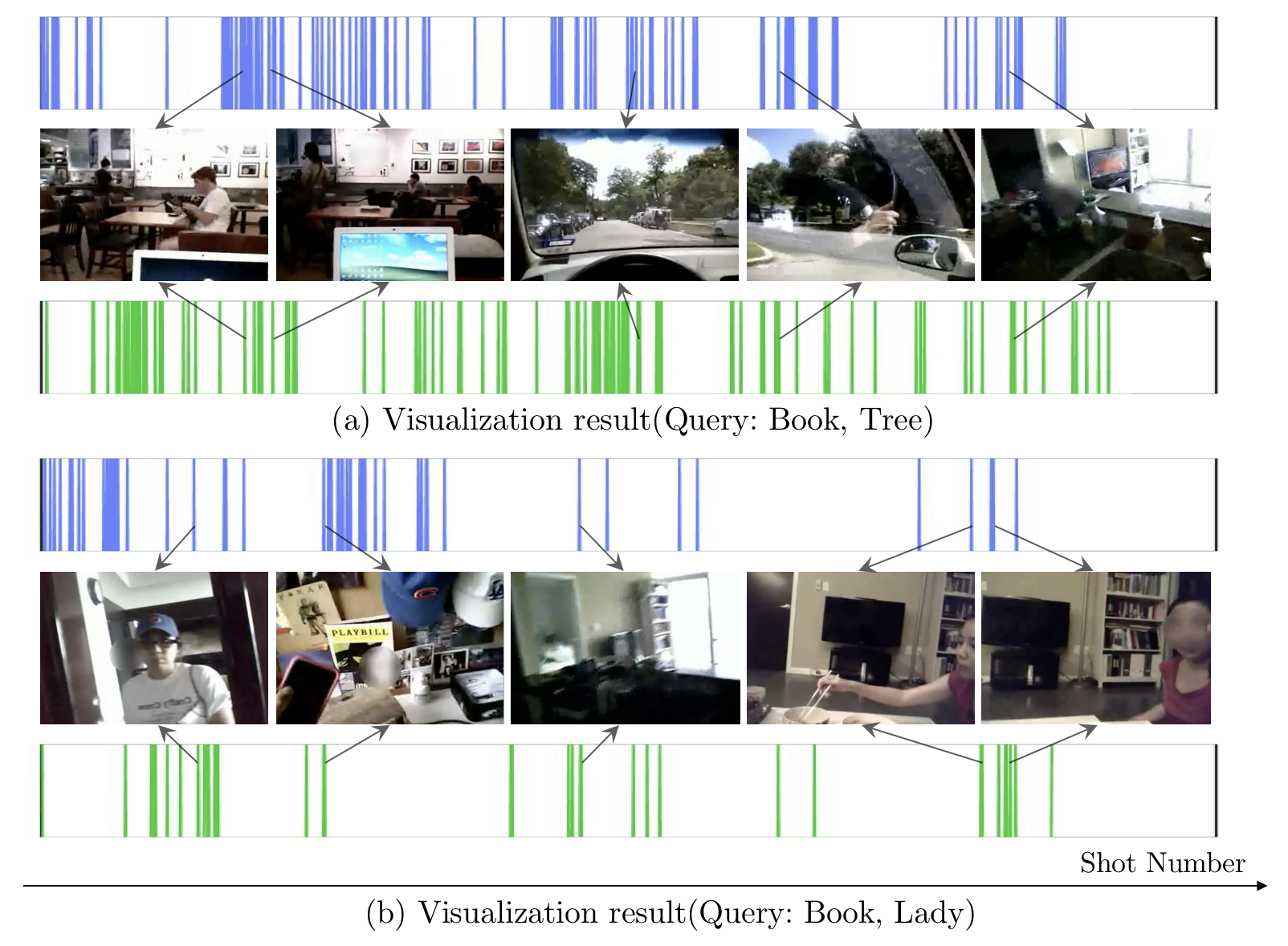}
    \caption{Some visualization results of our proposed method. The x-axis is the shot number given a certain video. The blue lines show the key shots of the ground-truths, and the green lines represent predicted key shots using our method. (a) The results for the query ``Book Tree''. (b) The results for the query ``Book Lady''.}
    \label{vis}
\end{figure*}

We provide some visualization results of our method in Figure.~\ref{vis}. We use the two user queries as examples: ``Book Tree'' and ``Book Lady'' (each user query contains two concepts). The x-axis in the figure is the shot number given a certain video. The upper blue lines denote the ground-truth key shots which are related to the user query, while the bottom green lines represent predicted key shots using our proposed method. Note that the selected summaries can be either related to one or two concepts given a user query. We can observe that our proposed method can find compact and representative summaries.

\section{Conclusions}
In this paper, we proposed a query-conditioned three-player generative adversarial network for query-conditioned video summarization. In the generator, video representations conditioned on user queries are obtained by jointly encoding visual information together with the text of user queries. Given these embeddings, confidence scores are predicted for each video shot in order to generate key shots based on these predicted scores. In the discriminator, we defined a three-player loss by introducing a randomly generated summary to prevent the model from generating trivial and short sequences. Experiments on videos of uncontrolled daily lives demonstrate the superiority of our proposed method.

\paragraph{Acknowledgment}
This project is supported by the Department of Defense under Contract No. FA8702-15-D-0002 with Carnegie Mellon University for the operation of the Software Engineering Institute, a federally funded research and development center. This work is also partially funded by the National Natural Science Foundation of China (Grant No. 61673378 and 61333016), and Norwegian Research Council FRIPRO grant no.\ 239844 on developing the \emph{Next Generation Learning Machines}.

\bibliography{main}
\end{document}